\pdfoutput=1

\documentclass[11pt]{article}

\usepackage[]{EMNLP2023}

\usepackage{times}
\usepackage{latexsym}
\usepackage{CJKutf8}

\usepackage{tikz}
\usetikzlibrary{fit}
\usetikzlibrary{positioning}
\usetikzlibrary{calc}
\usepackage[linguistics]{forest}

\usepackage[T1]{fontenc}

\usepackage[utf8]{inputenc}

\usepackage{microtype}

\usepackage{inconsolata}

\usepackage{todonotes}

\usepackage{listings,xcolor}

\lstdefinestyle{python}{
  language=Python,
  basicstyle=\fontsize{6.2}{5}\selectfont\ttfamily,
  keywordstyle=\color{magenta},
  stringstyle=\color{blue},
  commentstyle=\color{black!50}
}

\usepackage{bbm}
\usepackage{amsmath}

\newcommand{\jax}[1]{\texttt{jax.#1}}
\newcommand{\synjax}{\mbox{SynJax}}

\newcommand{\identity}[1]{\mathbbm{1}\!\!\left[#1\right]}

\newcommand{\figProj}{
    \begin{figure}[h]
        \centering
        \includegraphics[width=0.5\textwidth]{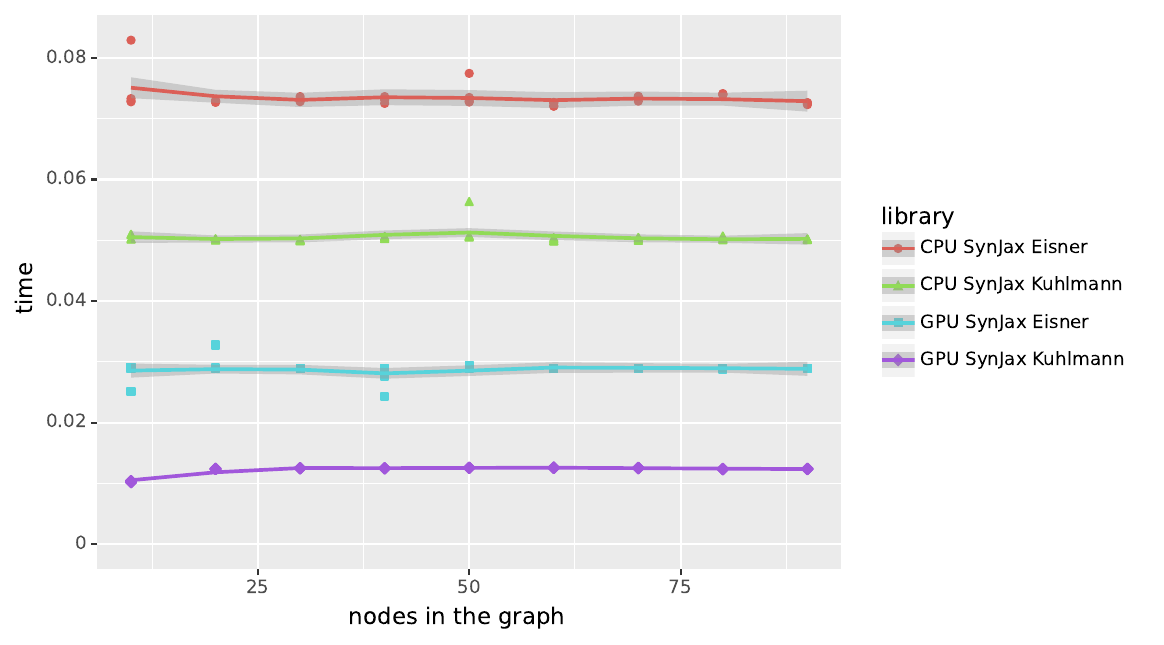}
        \caption{Speed comparison of Projective Maximum Spanning Tree algorithms.}
        \label{fig:projective}
    \end{figure}
}

\newcommand{\figNonProj}{
    \begin{figure}[h]
        \centering
        \includegraphics[width=0.5\textwidth]{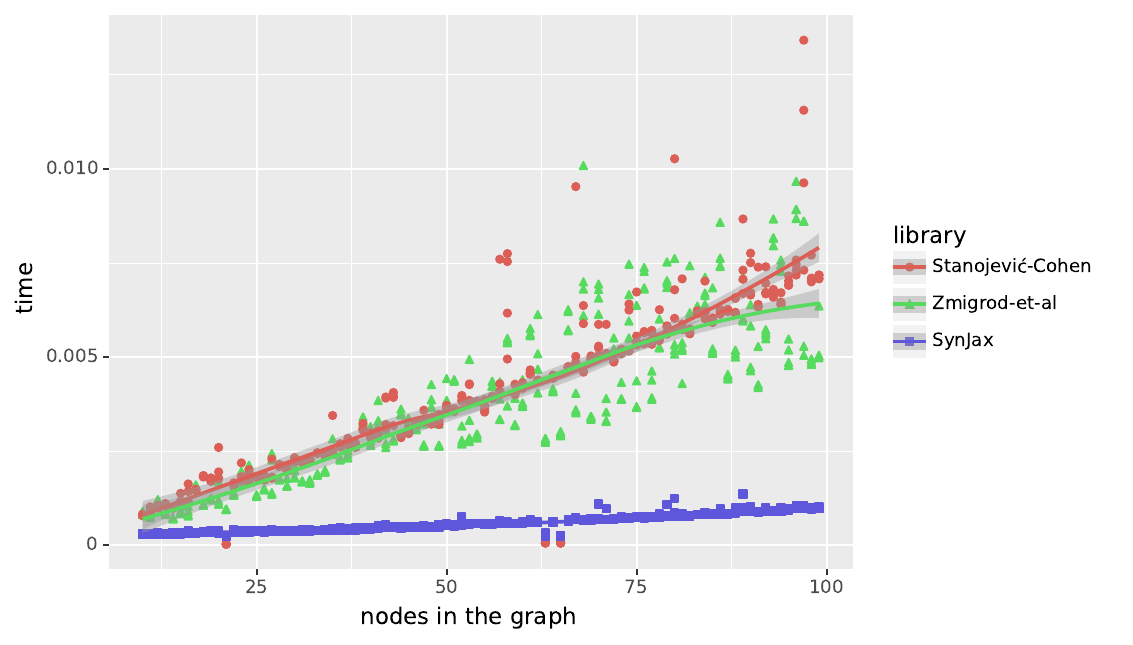}
        \caption{Speed comparison of Non-Projective Spanning Tree libraries.}
        \label{fig:non:projective}
    \end{figure}
}

\newcommand{\tableLinesOfCode}{
    \begin{table}[]
        \centering
        \scalebox{0.7}{
        \begin{tabular}{l|c|l|r}
                                  &        Torch-Struct  &        \ \ \ \ \ \ \ \ \synjax &  Speedup            \\ \hline
            Distribution          &            LoC       &        LoC \ \ (relative \%)   &                     \\\hline
            Linear-Chain-CRF      &            $32$      &         $15\hfill(46\%)$       &    $13\times$                \\
            Semi-Markov CRF       &            $54$      &         $15\hfill(27\%)$       &    $84\times$                \\
            Tree-CRF              &            $21$      &         $14\hfill(66\%)$       &    $ 5\times$                \\
            PCFG                  &            $51$      &         $36\hfill(70\%)$       &    $ 1\times$                \\
            Projective CRF        &            $70$      &         $54\hfill(77\%)$       &    $ 3\times$                \\
            Non-Projective CRF    &            $60$      &         $\ 8\hfill(16\%)$      &    $71\times$                \\
        \end{tabular}
        }
        \caption{Comparison against Torch-Struct with respect to lines of code for log-partition and relative speedup in the computation of marginal probabilities.}
        \label{table:loc:comparison}
    \end{table}
}

\newcommand{\figExampleTree}{
    \begin{figure}
        \centering
        \scalebox{0.64}{
            \begin{forest}
            for tree={s sep=10mm, inner sep=0, l=0}
             [S
                 [NP,tier=first
                     [\textcolor{red}{D}\\The,align=center,tier=words,name=A]
                     [\textcolor{red}{N}\\dog,align=center,tier=words,name=B] ]
                 [VP
                     [\textcolor{red}{V}\\chases,align=center,tier=words,name=C]
                     [NP,tier=first
                        [\textcolor{red}{D}\\a,align=center,tier=words,name=D]
                        [\textcolor{red}{N}\\cat,align=center,tier=words,name=E] ]
                 ]
             ]
             \node[draw,rounded rectangle,fit=(A) (B),fill=blue,opacity=.2] (AB) {};
             \node[draw,rounded rectangle,fit=(C),fill=blue,opacity=.2] (CC) {};
             \node[draw,rounded rectangle,fit=(D) (E),fill=blue,opacity=.2] (DE) {};
             \node [below =of D,draw,rounded rectangle,fill=yellow,opacity=.2,text opacity=1.] (CO) {\begin{CJK}{UTF8}{min} 追いかけている \end{CJK} };
             \node [left = 0.5cm of CO,draw,rounded rectangle,fill=yellow,opacity=.2,text opacity=1.] (BO) {\begin{CJK}{UTF8}{min} 猫を \end{CJK} };
             \node [left = 0.5cm of BO,draw,rounded rectangle,fill=yellow,opacity=.2,text opacity=1.] (AO) {\begin{CJK}{UTF8}{min} 犬が \end{CJK} };
             \draw[-=] (AB.south) to (AO.north);
             \draw[-=] (DE.south west) to[out=210,in=20] (BO.north east);
             \draw[-=] (CC.south) to (CO);
             \draw[->,dotted,red,very thick] ($(A.east)+(0,0.3)$) to ($(B.west)+(0,0.3)$);
             \draw[->,dotted,red,very thick] ($(B.east)+(0,0.3)$) to ($(C.west)+(0.3,0.3)$);
             \draw[->,dotted,red,very thick] ($(C.east)+(0,0.3)$) to ($(D.west)+(-0.1,0.3)$);
             \draw[->,dotted,red,very thick] ($(D.east)+(0.1,0.3)$) to ($(E.west)+(0,0.3)$);
            \end{forest}
        }
        \caption{Examples of natural language structures.}
        \label{fig:example:tree}
    \end{figure}
}

\title{\synjax{}: Structured Probability Distributions for JAX}

\author{Milo\v{s} Stanojevi\'{c} \\
  Google DeepMind \\
  \texttt{stanojevic@google.com} \\\And
  Laurent Sartran \\
  Google DeepMind \\
  \texttt{lsartran@google.com} \\}

\begin{document}
\maketitle
\begin{abstract}

The development of deep learning software libraries enabled significant progress in the field by allowing users to focus on modeling, while letting the library to take care of the tedious and time-consuming task of optimizing execution for modern hardware accelerators.
However, this has benefited only particular types of deep learning models, such as Transformers, whose primitives map easily to the vectorized computation.
The models that explicitly account for structured objects, such as trees and segmentations, did not benefit equally because they require custom algorithms that are difficult to implement in a vectorized form.

\synjax{} directly addresses this problem by providing an efficient vectorized implementation of inference algorithms for structured distributions covering alignment, tagging, segmentation, constituency trees and spanning trees.
This is done by exploiting the connection between algorithms for automatic differentiation and probabilistic inference.
With \synjax{} we can build large-scale differentiable models that explicitly model structure in the data.
The code is available at \mbox{\url{https://github.com/google-deepmind/synjax}}.
\end{abstract}

\section{Introduction}

In many domains, data can be seen as having some structure explaining how its parts fit into a larger whole. This structure is often latent, and it varies depending on the task. For examples of discrete structures in natural language consider Figure~\ref{fig:example:tree}. The words together form a sequence. Each word in a sequence is assigned a part-of-speech tag. These tags are dependent on each other, forming a linear-chain marked in red. 
The words in the sentence can be grouped together into small disjoint contiguous groups by sentence segmentation, shown with bubbles.
A deeper analysis of language would show that the groupings can be done recursively and thereby produce a syntactic tree structure.
Structures can also relate two languages. For instance, in the same figure, a Japanese translation can be mapped to an English source by an alignment.

These structures are not specific to language. Similar structures appear in biology as well. Nucleotides of any two RNA sequences are matched with monotone alignment \citep{needleman-wunsch-1970,rna:alignment:crf}, genomic data is segmented into contiguous groups \citep{genomic:segmentation:hmm} and tree-based models of RNA capture the hierarchical nature of the protein folding process
\citep{sakakibara1994stochastic,hockenmaier2007folding,liang:huang:rna}.

\figExampleTree{}

Most contemporary deep learning models attempt to predict output variables directly from the input without any explicit modeling of the intermediate structure. Modeling structure explicitly could improve these models in multiple ways. First, it could allow for better generalization trough the right inductive biases \citep{dyer-etal-2016-recurrent,sartran2022transformer}. This would improve not only sample efficiency but also downstream performance \citep{bastings-etal-2017-graph,nadejde-etal-2017-predicting,bisk-tran-2018-inducing}. Explicit modeling of structure can also enable incorporation of problem specific algorithms
\citep[e.g. finding shortest paths; ][]{Pogancic2020Differentiation,niepert2021implicit}
or constraints
(e.g. enforcing alignment \citeauthor{mena2018learning}, \citeyear{mena2018learning} or enforcing compositional calculation \citeauthor{havrylov-etal-2019-cooperative}, \citeyear{havrylov-etal-2019-cooperative}).
Discrete structure also allows for better interpretability of the model's decisions \citep{bastings2019interpretable}. Finally, sometimes structure is the end goal of learning itself -- for example we may know that there is a hidden structure of a particular form explaining the data, but its specifics are not known and need to be discovered \citep{kim-etal-2019-compound,paulus-et-al-2020-neurips}.

Auto-regressive models are the main approach used for modeling sequences. Non-sequential structures are sometimes linearized and approximated with a sequential structure \citep{choe-charniak-2016-parsing}. These models are powerful as they do not make any independence assumptions and can be trained on large amounts of data. While sampling from auto-regressive models is typically tractable, other common inference problems like finding the optimal structure or marginalizing over hidden variables are not tractable.
Approximately solving these tasks with auto-regressive models requires using biased or high-variance approximations that are often computationally expensive, making them difficult to deploy in large-scale models.

Alternative to auto-regressive models are models over factor graphs that factorize in the same way as the target structure. These models can efficiently compute all inference problems of interest exactly by using specialized algorithms. Despite the fact that each structure needs a different algorithm, we do not need a specialized algorithm for each inference task (argmax, sampling, marginals, entropy etc.). As we will show later, \synjax{} uses automatic differentiation to derive many quantities from just a single function per structure type.

Large-scale deep learning has been enabled by easy to use libraries that run on hardware accelerators. Research into structured distributions for deep learning has been held back by the lack of ergonomic libraries that would provide accelerator-friendly implementations of structure components -- especially since these components depend on algorithms that often do not map directly onto available deep learning primitives, unlike Transformer models. This is the problem that \synjax{} addresses by providing easy to use structure primitives that compose within JAX machine learning framework.

To see how easy it is to use \synjax{} consider example in Figure~\ref{fig:example:code}. This code implements a policy gradient loss that requires computing multiple quantities -- sampling, argmax, entropy, log-probability -- each requiring a different algorithm. In this concrete code snippet, the structure is a non-projective directed spanning tree with a single root edge constraint. Because of that \synjax{} will:
\begin{itemize}
\setlength{\itemsep}{1pt}
\setlength{\parskip}{0pt}
\setlength{\parsep}{0pt}
\item compute argmax with Tarjan's (\citeyear{tarjan77}) maximum spanning tree algorithm adapted for single root edge trees \citep{stanojevic-cohen-2021-emnlp},
\item sample with Wilson's (\citeyear{wilson}) sampling algorithm for single root trees \citep{stanojevic-2022-unbiased},
\item compute entropy with Matrix-Tree Theorem \citep{Tutte84} adapted for single root edge trees \citep{koo-etal-2007-structured,zmigrod:expectation}.
\end{itemize}

If the user wants only to change slightly the the tree requirements to follow the \emph{projectivity constraint} they only need to change one flag and \synjax{} will in the background use completely different algorithms that are appropriate for that structure: it will use Kuhlmann's algorithm (\citeyear{kuhlmann-etal-2011-dynamic}) for argmax and variations of Eisner's (\citeyear{eisner-1996-three}) algorithm for other quantities.
The user does not need to implement any of those algorithms or even be aware of their specifics, and can focus on the modeling side of the problem.

\begin{figure}
    \centering
    \begin{lstlisting}[
    style=python
]
@typed
def policy_gradient_loss(
    log_potentials: Float[jax.Array, "*batch n n"],
    key: jax.random.KeyArray) -> Float[jax.Array, ""]:
  dist = synjax.SpanningTreeCRF(log_potentials,
    directed=True, projective=False, single_root_edge=True)
  # Sample from policy
  sample = dist.sample(key)
  # Get reward
  reward = reward_fn(sample)
  # Compute log-prob
  log_prob = dist.log_prob(sample)
  # Self-critical baseline
  baseline = reward_fn(dist.argmax())
  # REINFORCE
  objective = stop_gradient(reward-baseline) * log_prob
  # Entropy regularization
  return -jnp.mean(objective + 0.5*dist.entropy())
    \end{lstlisting}
    \caption{Example of implementing policy gradient with self-critical baseline and entropy regularization for spanning trees.}
    \label{fig:example:code}
\end{figure}

\section{Structured Distributions}

Distributions over most structures can be expressed with factor graphs -- bipartite graphs that have random variables and factors between them.
We associate to each factor a non-negative scalar, called potential, for each possible assignment of the random variables that are in its neighbourhood. The potential of the structure is a product of its factors:

\begin{equation}
    \phi(t) = \prod_{e \in t} \phi(e)
\end{equation}
where $t$ is a structure, $e$ is a factor/part, and $\phi(\cdot)$ is the potential function.
The probability of a structure can be found by normalizing its potential:
\begin{equation}
    p(t) = \frac{\prod_{e \in t} \phi(e)}{\sum_{t' \in T} \prod_{e' \in t'} \phi(e') } = \frac{\phi(t)}{Z} \label{eq:glob:prob}
\end{equation}
where $T$ is the set of all possible structures and $Z$ is a normalization often called partition function.
This equation can be thought of as a \emph{softmax} equivalent over an extremely large set of structured outputs that share sub-structures \citep{sutton20074an,mihaylova-etal-2020-understanding}.

\section{Computing Probability of a Structure and Partition Function}

Equation \ref{eq:glob:prob} shows the definition of the probability of a structure in a factor graph. Computing the numerator is often trivial. However, computing the denominator, the partition function, is the complicated and computationally demanding part because the set of valid structures $T$ is usually exponentially large and require specialized algorithms for each type of structure. As we will see later, the algorithm for implementing the partition function accounts for the majority of the code needed to add support for a structured distribution, as most of the other properties can be derived from it. Here we document the algorithms for each structure.

\subsection{Sequence Tagging}

\newcommand{\linearChainCRF}{Linear-Chain CRF}

Sequence tagging can be modelled with \linearChainCRF{} \citep{crf:linear:chain}. The partition function for linear-chain models is computed with the forward algorithm \citep{rabiner1990tutorial}.
The computational complexity is $\mathcal{O}(m^2n)$ for $m$ tags and sequence of length $n$.
\citet{sarkka-garcia-fernandez-2019} have proposed a parallel version of this algorithm that has parallel computational complexity $\mathcal{O}(m^3\log n)$ which is efficient for $m\!\!\ll\!\!n$.
\citet{torch-struct} reports a speedup using this parallel method for Torch-Struct, however in our case the original forward algorithm gave better performance both in terms of speed and memory. %

The \synjax{} implementation of \linearChainCRF{} supports having a different transition matrix for each time step which gives greater flexibility needed for implementing models like \mbox{LSTM-CNN-CRF} \citep{ma-hovy-2016-end} and Neural Hidden Markov Model \citep{tran-etal-2016-unsupervised-neural-hmm}.

\subsection{Segmentation with Semi-Markov CRF}

Joint segmentation and tagging can be done with a generalization of linear-chain called Semi-Markov CRF \citep{semi:markov:crf:sarawagi:cohen,segmental:hamid:2013,lu2016segmental}. It has a similar parametrization with transition matrices except that here transitions can jump over multiple tokens.
The partition function is computed with an adjusted version of the forward algorithm that runs in $\mathcal{O}(sm^2n)$ where $s$ is the maximal size of a segment.

\subsection{Alignment Distributions}

Alignment distributions are used in time series analysis \citep{soft-dtw:2017}, RNA sequence alignment \citep{rna:alignment:crf}, semantic parsing \citep{lyu-titov-2018-amr} and many other areas.

\subsubsection{Monotone Alignment}

Monotone alignment between two sequences of lengths $n$ and $m$ allows for a tractable partition function that can be computed in $\mathcal{O}(nm)$ time using the Needleman-Wunsch (\citeyear{needleman-wunsch-1970}) algorithm.

\subsubsection{CTC}

Connectionist Temporal Classification \citep[CTC,][]{graves2006connectionist,hannun2017sequence} is a monotone alignment model widely used for speech recognition and non-auto-regressive machine translation models.
It is distinct from the standard monotone alignment because it requires special treatment of the \emph{blank symbol} that provides jumps in the alignment table. It is implemented with an adjusted version of Needleman-Wunsch algorithm.

\subsubsection{Non-Monotone 1-on-1 Alignment}

This is a bijective alignment that directly maps elements between two sets given their matching score. Computing partition function for this distribution is intractable \citep{VALIANT:intractable}, but we can compute some other useful quantities (see Section~\ref{sec:argmax}).

\subsection{Constituency Trees}

\subsubsection{Tree-CRF}

Today's most popular constituency parser by \citet{kitaev-etal-2019-multilingual} uses a global model with factors defined over labelled spans. \citet{stern-etal-2017-minimal} have shown that inference in this model can be done efficiently with a custom version of the CKY algorithm in $\mathcal{O}(mn^2+n^3)$ where $m$ is number of non-terminals and $n$ is the sentence length.

\subsubsection{PCFG}

Probabilistic Context-Free Grammars (PCFG) are a generative model over constituency trees where each grammar rule is associated with a locally normalized probability. These rules serve as a template which, when it gets expanded, generates jointly a constituency tree together with words as leaves.

\synjax{} computes the partition function using a vectorized form of the CKY algorithm that runs in cubic time.
Computing a probability of a tree is in principle simple: just enumerate the rules of the tree, look up their probability in the grammar and multiply the found probabilities. However, extracting rules from the set of labelled spans requires many sparse operations that are non-trivial to vectorize. We use an alternative approach where we use \emph{sticky} span log-potentials to serve as a mask for each constituent: constituents that are part of the tree have sticky log-potentials $0$ while those that are not are $-\infty$. With sticky log-potentials set in this way computing log-partition provides a log-probability of a tree of interest.

\subsubsection{TD-PCFG}

Tensor-Decomposition PCFG \citep[TD-PCFG,][]{cohen-etal-2013-approximate,yang-etal-2022} uses a lower rank tensor approximation of PCFG that makes inference with much larger number of non-terminals feasible.

\subsection{Spanning Trees}

Spanning trees appear in the literature in many different forms and definitions. We take a spanning tree to be any subgraph that connects all nodes and does not have cycles. We divide spanning tree CRF distributions by the following three properties:
\setlength{\parskip}{0pt}
\begin{description}
\setlength{\itemsep}{1pt}
\setlength{\parskip}{0pt}
\setlength{\parsep}{0pt}
\item[directed or undirected] Undirected spanning trees are defined over symmetric weighted adjacency matrices i.e. over undirected graphs. Directed spanning trees are defined over directed graphs with special root node.
\item[projective or non-projective] Projectivity is a constraint that appears often in NLP. It constrains the spanning tree over words not to have crossing edges. Non-projective spanning tree is just a regular spanning tree -- i.e. it may not satisfy the projectivity constraint.
\item[single root edge or multi root edges] NLP applications usually require that there can be only one edge coming out of the root \citep{zmigrod:please:mind:the:root}. Single root edge spanning trees satisfy that constraint.
\end{description}

Each of these choices has direct consequences on which algorithm should be used for probabilistic inference. \synjax{} abstracts away this from the user and offers a unified interface where the user only needs to provide the weighted adjacency matrix and set the three mentioned boolean values. Given the three booleans \synjax{} can pick the correct and most optimal algorithm.
In total, these parameters define distributions over 8 different types of spanning tree structures all unified in the same interface. We are not aware of any other library providing this set of unified features for spanning trees.

We reduce undirected case to the rooted directed case due to bijection.
For projective rooted directed spanning trees we use Eisner's algorithm for computation of the partition function \citep{eisner-1996-three}. The partition function of Non-Projective spanning trees is computed using Matrix-Tree Theorem
\citep{Tutte84,koo-etal-2007-structured,smith-smith-2007-probabilistic}.

\section{Computing Marginals}

In many cases we would like to know the probability of a particular part of structure appearing, regardless of the structure that contains it. In other words, we want to marginalize (i.e. sum) the probability of all the structures that contain that part:
\begin{equation}
    p(e) = \sum_{t \in T} \identity{e \in t}\ p(t) = \sum_{t' \in T_e} p(t')
\end{equation}
where $\identity{\cdot}$ is the indicator function, $T$ is the set of all structures and $T_e$ is the set of structures that contain factor/part $e$.

Computing these factors was usually done using specialized algorithms such as Inside-Outside or Forward-Backward. However, those solutions do not work on distributions that cannot use belief propagation like Non-Projective Spanning Trees. A more general solution is to use an identity that relates gradients of factor's potentials with respect to the log-partition function:
\begin{equation}
    p(e) = \frac{\partial \log Z}{\partial \phi(e)} \label{eq:derivative:trick}
\end{equation}

This means that we can use any differentiable implementation of log-partition function as a forward pass and apply backpropagation to compute the marginal probability \citep{darwiche:2003}. \citet{eisner-2016-inside} has made an explicit connection that ``Inside-Outside and Forward-Backward algorithms are just backprop''. This approach also works for Non-Projective Spanning Trees that do not fit belief propagation framework \citep{zmigrod:expectation}.

For template models like PCFG, we use again the \emph{sticky} log-potentials because usually we are not interested in marginal probability of the rules but in the marginal probability of the instantiated constituents.
The derivative of log-partition with respect to the constituent's \emph{sticky} log-potential will give us marginal probability of that constituent.

\section{Computing Most Probable Structure}
\label{sec:argmax}

For finding the score of the highest scoring structure we can just run the same belief propagation algorithm for log-partition, but with the \emph{max-plus semiring} instead of the log-plus semiring \citep{goodman-1999-semiring}. To get the most probable structure, and not just its potential, we can compute the gradient of part potentials with respect to the viterbi structure potential \citep{torch-struct}.

The only exceptions to this process are non-monotone alignments and spanning trees because they do fit easily in belief propagation framework.
For the highest scoring non-monotone alignment, we use the Jonker–Volgenant algorithm as implemented in SciPy \citep{crouse2016implementing,2020SciPy-NMeth}. Maximal \emph{projective} spanning tree can be found by combining Eisner's algorithm with max-plus semiring, but we have found Kuhlmann's tabulated arc-hybrid algorithm to be much faster \citep{kuhlmann-etal-2011-dynamic} (see Figure~\ref{fig:projective} in the appendix). This algorithm cannot be used for any inference task other than argmax because it allows for spurious derivations. To enforce single-root constraint with Kuhlmann's algorithm we use the Reweighting trick from \citet{stanojevic-cohen-2021-emnlp}. For \emph{non-projective} spanning trees \synjax{} uses a combination of Reweighting+Tarjan algorithm as proposed in \citet{stanojevic-cohen-2021-emnlp}.

\section{Sampling a Structure}

Strictly speaking, there is no proper sampling semiring because semirings cannot have non-deterministic output. However, we can still use the semiring framework and make some aspect of them non-deterministic.
\citet{wilker:grasp} and \citet{torch-struct} use a semiring that in the forward pass behaves like a log-semiring, but in the backward pass instead of computing the gradient it does sampling. This is in line of how forward-filtering backward-sampling algorithm works \citep[\S17.4.5]{Murphy:2012}.

Non-Projective Spanning Trees do not support the semiring framework so we use custom algorithms for them described in \citet{stanojevic-2022-unbiased}. It contains Colbourn's algorithm that has a fixed runtime of $\mathcal{O}(n^3)$ but is prone to numerical issues because it requires matrix-inversion \citep{colbourn}, and Wilson's algorithm that is more numerically stable but has a runtime that depends on concrete values of log-potentials \citep{wilson}.
\synjax{} also supports vectorized sampling without replacement (SWOR) from \citet{stanojevic-2022-unbiased}.

\section{Differentiable Sampling}

The mentioned sampling algorithms provide unbiased samples of structures useful for many inference tasks, but they are not differentiable because the gradient of sampling from discrete distributions is zero almost everywhere.
This problem can be addressed with log-derivative trick from REINFORCE algorithm \citep{reinforce:williams}, but that provides high variance estimates of gradients.
To address this problem there have been different proposals for differentiable sampling algorithms that are biased but can provide low-variance estimates of gradients.
\synjax{} implements majority of the main approaches in the literature including
structured attention \citep{kim2017structured},
relaxed dynamic programming \citep{mensch-blondel-2018},
Perturb-and-MAP \citep{corro2018differentiable},
Gumbel-CRF \citep{fu-2020-gumbel-crf},
Stochastic Softmax-Tricks \citep{paulus-et-al-2020-neurips},
and Implicit Maximum-Likelihood estimation \citep{niepert2021implicit}.
It also include different noise distributions for perturbations models, including Sum-of-Gamma noise \citep{niepert2021implicit} that is particularly suited for structured distributions.

\section{Entropy and KL Divergence}

To compute the cross-entropy and KL divergence, we will assume that the two distributions factorize in exactly the same way. Like some other properties, cross-entropy can also be computed with the appropriate semirings \citep{hwa-2000-sample,eisner-2002-parameter-entropy-semiring,cortes:entropy:2008,log:entropy:semiring:2023}, but those approaches would not work on Non-Projective Spanning Tree distributions. There is a surprisingly simple solution that works across all distributions that factorize in the same way and has appeared in a couple of works in the past \citep{li-eisner-2009-first,martins-etal-2010-turbo,zmigrod:expectation}. Here we give a full derivation for cross-entropy:
\begin{align}
 H(p, q) & = - \sum_{t \in T} p(t) \log q(t) \nonumber \\
      & = \log Z_q - \sum_{t \in T} p(t) \sum_{e \in t} \log \phi_q(e) \nonumber \\
      & = \log Z_q - \sum_{t \in T} p(t) \sum_{e \in E} \identity{e\!\in\!t} \log \phi_q(e) \nonumber \\
      & = \log Z_q - \sum_{e \in E} p(e) \log \phi_q(e) \label{eq:cross:entropy} %
\end{align}
This reduces the computation of cross-entropy to finding marginal probabilities of one distribution, and finding log-partition of the other -- both of which can be computed efficiently for all distributions in \synjax{}. Given the method for computing cross-entropy, finding entropy is trivial:
\begin{equation}
 H(p) = H(p, p) \label{eq:entropy}
\end{equation}

KL divergence is easy to compute too:
\begin{equation}
 D_{KL}(p || q) = H(p, q) - H(p) \label{eq:kl:divergence}
\end{equation}

\section{Library Design}

Each distribution has different complex shape constraints which makes it complicated to document and implement all the checks that verify that the user has provided the right arguments.
The \mbox{\emph{jaxtyping}} library\footnote{\url{https://github.com/google/jaxtyping}} was very valuable in making \synjax{} code concise, documented and automatically checked. 

Structured algorithms require complex broadcasting, reshaping operations and application of semirings. To make this code simple, we took the \emph{einsum} implementation from the core JAX code and modified it to support arbitrary semirings. This made the code shorter and easier to read.

Most inference algorithms apply a large number of elementwise and reshaping operations that are in general fast but create a large number of intermediate tensors that occupy memory. To speed this up we use checkpointing \citep{checkpointing} to avoid memorization of tensors that can be recomputed quickly.
That has improved memory usage \emph{and} speed, especially on TPUs.

All functions that could be vectorized are written in pure JAX.
Those that cannot, like Wilson sampling (\citeyear{wilson}) and Tarjan's algorithm (\citeyear{tarjan77}), are implemented with Numba \citep{numba}.

All \synjax{} distributions inherit from Equinox modules \citep{equinox} which makes them simultaneously PyTrees and dataclasses.
Thereby all \synjax{} distributions can be transformed with \jax{vmap} and are compatible with any JAX neural framework, e.g. Haiku and Flax.

\section{Comparison to alternative libraries}
\label{sec:alternatives}

JAX has a couple of libraries for probabilistic modeling.
Distrax \citep{deepmind2020jax} and Tensorflow-Probability JAX substrate \citep{tensorflow:probability:distributions:2017} provide continuous distributions.
NumPyro \citep{phan2019composable:numpyro} and Oryx provide probabilistic programming.
DynaMax \citep{dynamax:github} brings state space models to JAX and includes an implementation of HMMs.

\tableLinesOfCode{}

PGMax \citep{zhou2023pgmax} is a JAX library that supports inference over arbitrary factor graphs by using loopy belief propagation.
After the user builds the desired factor graph, \mbox{PGMax} can do automatic inference over it.
For many structured distributions building a factor graph is the  difficult part of implementation because it may require a custom algorithm (e.g. CKY or Needleman–Wunsch).
\synjax{} implements those custom algorithms for each of the supported structures. With \synjax{} the user only needs to provide the parameters of the distribution and \synjax{} will handle \emph{both} building of the factor graph and inference over it.
For all the included distributions, \synjax{} also provides some features not covered by \mbox{PGMax}, such as unbiased sampling, computation of entropy, \mbox{cross-entropy} and KL divergence.

Optax \citep{deepmind2020jax} provides CTC loss implementation for JAX but without support for computation of optimal alignment, marginals over alignment links, sampling alignments etc.

All the mentioned JAX libraries focus on continuous or categorical distributions and, with the exception of HMMs and CTC loss, do not contain distributions provided by \synjax{}.
\synjax{} fills this gap in the JAX ecosystem and enables easier construction of structured probability models.

The most comparable library in terms of features is Torch-Struct \citep{torch-struct} that targets \mbox{PyTorch} as its underlying framework.
\mbox{Torch-Struct}, just like \synjax{}, uses automatic differentiation for efficient inference.
We will point out here only the main differences that would be of relevance to users.
\synjax{} supports larger number of distributions and inference algorithms and provides a unified interface to all of them.
It also provides reproducable sampling trough controlled randomness seeds.
\synjax{} has a more general approach to computation of entropy that does not depend on semirings and therefore applies to all distributions.
\synjax{} is fully implemented in Python and compiled with \jax{jit} and \texttt{numba.jit} while Torch-Struct does not use any compiler optimizations except a custom CUDA kernel for semiring matrix multiplication.
If we compare lines of code and speed (Table~\ref{table:loc:comparison}) we can see that \synjax{} is much more concise and faster than Torch-Struct (see Appendix~\ref{appendix:empirical} for details).

\synjax{} also provides the fastest and most feature rich implementation of spanning tree algorithms. So far the most competitive libraries for spanning trees were by \citeauthor{zmigrod:expectation} and \citeauthor{stanojevic-cohen-2021-emnlp}. \synjax{} builds on \citeauthor{stanojevic-cohen-2021-emnlp} code and annotates it with Numba instructions which makes it many times faster than any other alternative (see Figure~\ref{fig:non:projective} in the appendix).

\section{Conclusion}

One of the main challenges in creating deep neural models over structured distributions is the difficulty of their implementation on modern hardware accelerators. \synjax{} addresses this problem and makes large scale training of structured models feasible and easy in JAX.
We hope that this will encourage research into finding alternatives to auto-regressive modeling of structured data.

\section*{Limitations}
\synjax{} is quite fast, but there are still some areas where the improvements could be made.
One of the main speed and memory bottlenecks is usage of big temporary tensors in the dynamic programming algorithms needed for computation of log-partition function. This could be optimized with custom kernels written in 
Pallas.\footnote{\url{https://jax.readthedocs.io/en/latest/pallas}}
There are some speed gains that would conceptually be simple but they depend on having a specialized hardware. For instance, matrix multiplication with semirings currently does not use hardware acceleration for matrix multiplication, such as TensorCore on GPU, but instead does calculation with regular CUDA cores.
We have tried to address this with log-einsum-exp trick \citep{log:einsum:exp:trick} but the resulting computation was less numerically precise than using a regular log-semiring with broadcasting.
Maximum spanning tree algorithm would be much faster if it could be vectorized -- currently it's executing as an optimized Numba CPU code.

\section*{Acknowledgements}
We are grateful to Chris Dyer, Aida Nematzadeh and other members of language team in Google DeepMind for early comments on the draft of this work.
We appreciate Patrick Kidger's work on Equinox and Jaxtyping that made development of \synjax{} much easier.
We also appreciate that Sasha Rush open-sourced Torch-Struct, a library that influenced many aspects of \synjax{}.

\bibliography{BIB}
\bibliographystyle{acl_natbib}

\appendix

\section{Empirical comparisons}
\label{appendix:empirical}

\subsection{Comparison with Torch-Struct}

We compared against the most recent Torch-Struct\footnote{\url{https://github.com/harvardnlp/pytorch-struct}} commit from 30 Jan 2022. To make Torch-Struct run faster we have also installed its specialized kernel for semiring matrix multiplication \emph{genbmm}\footnote{\url{https://github.com/harvardnlp/genbmm}} from its most recent commit from 11 Oct 2021. While Torch-Struct supports some of the same distributions as \synjax{} we did not manage to do speed comparison over all of them. For example, AlignmentCRF of Torch-Struct was crashing due to mismatch of PyTorch, Torch-Struct and genbmm changes about in-place updates.
We compile \synjax{} with \jax{jit} and during benchmarking do not count the time that is taken for compilation because it needs to be done only once. We also tried to compile Torch-Struct using TorchScript by tracing but that did not work out of the box.
Comparisons are done on A100 GPU on Colab Pro+. The results are shown in Table~\ref{table:loc:comparison} in the main text.   Table~\ref{tab:distribution:params} shows sizes of the distributions being tested.

\begin{table}[]
    \centering
    \scalebox{0.85}{
        \begin{tabular}{l|l}
          Distribution       & parameters                       \\ \hline\hline
          CTC                & b=\phantom{00}64,\ \ n=124,\ \ l=512               \\
          Alignment CRF      & b=\phantom{00}16,\ \ n=256,\ \ m=256               \\
          Semi-Markov CRF    & b=\phantom{000}1,\ \ n=\phantom{0}64,\ \ nt=\phantom{0}32, k=8            \\
          Tree CRF           & b=\phantom{0}128,\ \ n=128,\ \ nt=128             \\
          \linearChainCRF{}  & b=\phantom{0}128,\ \ n=256,\ \ nt=\phantom{0}32              \\
          PCFG               & b=\phantom{000}1,\ \ n=\phantom{0}48,\ \ nt=\phantom{0}64,\ \ pt=96          \\
          HMM                & b=\phantom{000}1,\ \ n=128,\ \ nt=\phantom{0}32                \\
          Non-Projective CRF & b=\phantom{}1024,\ \ n=128                    \\
          Projective CRF     & b=\phantom{0}128,\ \ n=128                     \\
        \end{tabular}
    }
    \caption{Sizes of tested distributions.}
    \label{tab:distribution:params}
\end{table}

\subsection{Comparison with \citeauthor{zmigrod:please:mind:the:root}}

Non-Projective spanning trees present a particular challenge because they cannot be vectorized easily due to dynamic structures that are involved in the algorithm. The main algorithms and libraries for parsing this type of trees are from \citet{zmigrod:please:mind:the:root}\footnote{\url{https://github.com/rycolab/spanningtrees}}
and \citet{stanojevic-cohen-2021-emnlp}\footnote{\url{https://github.com/stanojevic/Fast-MST-Algorithm}}. The first one is expressed as a recursive algorithm, while the second one operates over arrays of fixed size in iterative way. This makes \citeauthor{stanojevic-cohen-2021-emnlp} algorithm much more amendable to Numba optimization. We took that code and just annotated it with Numba primitives. This made the algorithm significantly faster, especially for big graphs, as can be seen from Figure~\ref{fig:non:projective}.

\figNonProj{}

\subsection{Comparison of Maximum Projective Spanning Tree Algorithms}

Eisner's algorithm is virtually the only projective parsing algorithm actively used, if we do not count the transition based parsers. We have found that replacing Eisner's algorithm with \citet{kuhlmann-etal-2011-dynamic} tabulation of arc-hybrid algorithm can provide large speed gains both on CPU and GPU. See Figure~\ref{fig:projective}. In this implementation graph size does not make a big difference because it is implemented in a vectorized way so most operations are parallelized.

\figProj{}

\end{document}